\documentclass[]{article}
\usepackage{proceed2e}

\usepackage{times}
\usepackage{graphicx} 
\usepackage{subfigure} 

\usepackage{hyperref}

\usepackage{qiangstyle}
\usepackage{times}

\title{Learning Deep Energy Models: Contrastive Divergence vs. Amortized MLE}

\author{Qiang Liu  ~~~~~~~~~~~~~~~~  Dilin Wang \\
Computer Science, Dartmouth College, Hanover, NH 03755}

\begin{document}

\maketitle


\begin{abstract} 
We propose a number of new algorithms for learning deep energy models from data motivated by a recent Stein variational gradient descent  (SVGD) algorithm, 
including a Stein contrastive divergence (SteinCD) that integrates CD with SVGD based on their theoretical connections, 
and a SteinGAN that 
trains an auxiliary generator 
to generate the negative samples in maximum likelihood estimation (MLE). 
We demonstrate that our SteinCD trains models with good generalization (high test likelihood), 
while SteinGAN can generate realistic looking images competitive with GAN-style methods. 
We show that by combing SteinCD and SteinGAN, it is possible to inherent the advantage of both approaches. 
\end{abstract} 

\section{Introduction}
Energy-based models (EBMs) capture dependencies between variables by associating a scalar energy to each configuration of the variables.
Learning EBMs consists in finding an energy function that assigns low energy to correct values, and high energy to
incorrect values. 
Energy-based learning provides a unified framework for many learning models,
such as undirected graphical models \citep{lecun2006tutorial}, deep generative models \citep{ngiam2011learning,xie2016theory}. 

Maximum likelihood estimator (MLE) provides a fundamental approach for learning energy-based probabilistic models from data. Unfortunately, exact MLE is intractable to calculate due to the difficulty of evaluating the normalization constant and its gradient. This problem has attracted a vast literature in the last few decades, based on either approximating the likelihood objective, or developing alternative surrogate loss functions \citep[see e.g.,][for reviews]{koller2009probabilistic, goodfellow2016deep}. 
Contrastive divergence (CD) \citep{hinton2002training} is one of the most important algorithms,  
which avoids estimating the normalization constant by optimizing a 
contrastive objective that measures how much KL divergence can be improved by 
running a small numbers of Markov chain steps towards the intractable energy model. 
CD has been widely used for learning models like restricted Boltzmann machines and Markov random fields \citep{carreira2005contrastive,hinton2006reducing}. 

Although being able to train models that have high testing likelihood, 
CD and other traditional energy-based learning algorithms can not generate high quality samples that resemble real-world instances, such as realistic-looking images. 
This is because the real world instances live a relatively low manifold which the the energy-based models can not capture. 
This problem has been addressed by the recent generative adversarial networks (GAN) \citep[e.g.,][to name only a few]{goodfellow2014generative,radford2015unsupervised,salimans2016improved, wgan}, 
which, instead of training energy models, 
directly train generative networks that output random samples 
to match the observed data 
by framing the divergence minimization problem into a minimax game. 
By designing the generator using deep convolutional networks \citep{radford2015unsupervised}, 
the prior knowledge of the real-world manifold can be incorporated into learning. 
However, GAN does not explicitly assign an energy score for each data point, 
and can over-fit on a subset of the training data, and ignore the remaining ones. 
A promising direction is to combine GAN-type methods with traditional energy-based learning to integrate the advantages of both. 

Based on a recent Stein variational gradient descent (SVGD) algorithm for approximate inference \citep{liu2016stein},  
we propose a number of new algorithms for training deep energy models, 
including a \emph{Stein contrastive divergence} (SteinCD) that combines CD with SVGD based on their theoretical connections,  
and a SteinGAN algorithm 
that approximates MLE using a sampler (generator) that amortizes the negative sample approximation. 
We show that SteinCD and SteinGAN exhibit opposite properties, 
SteinCD tends to learn models with high testing likelihood but can not generate high quality images, 
while SteinGAN generates realistic looking images but does not generalize well. 
Our SteinGAN approach suggests that it is possible to generate high quality images comparable with GAN-type methods using energy-based models, 
opening the possibility of combining the traditional energy-based learning techniques with GAN approaches. 
In experiments, we show evidence that by simply mixing SteinCD and SteinGAN updates it is possible to obtain algorithms that combine the advantage of both. 
\paragraph{Outline} Section~\ref{sec:backgroud} introduces background on Stein variational gradient descent and energy-based models. 
Section~\ref{sec:steincd} and \ref{sec:steingan} discuss our SteinCD and SteinGAN methods for training energy-based models, respectively. 
Empirical results are shown in Section~\ref{sec:experiments}. 
Section \ref{sec:conclusion} concludes the paper. 

\section{Background}
\label{sec:backgroud}
In this section, 
we first introduce the background of Stein variational gradient descent (SVGD) which forms the foundation of our work,  
and then review energy-based probabilistic models and contrastive divergence (CD). 
Our introduction highlights the connection between SVGD and CD which motivates us to propose SteinCD in Section~\ref{sec:steincd}.

\subsection{Stein Variational Gradient Descent (SVGD)}
\label{sec:svgd}

\newcommand\eqdef{\stackrel{\mathclap{\scriptsize\mbox{def}}}{=}}
Stein variational gradient descent (SVGD) \citep{liu2016stein} is a general purpose deterministic approximate sampling method.
The idea is to iteratively evolve a set of particles to yield the fastest decrease of KL divergence locally. 

Let $p(x)$ be a positive density function in $\RR^d$ that we want to approximate. 
Assume we start with a set of particles $\{x_i\}_{i=1}^n$ whose empirical distribution is $q_0(x) = \sum_i \delta(x - x_i)/n$, 
and want to move $\{x_i\}_{i=1}^n$ closer to the target distribution $p(x)$ to improve the approximation quality. 
To do so, assume we update the particles by a transform of form
\vspace{-3pt}
$$
x_i' \leftarrow x_i + \epsilon \ff(x_i), ~~~\forall i = 1, \cdots, n,
$$
where $\epsilon$ is a small step size and $\ff$ is a velocity field that decides the perturbation direction of the particles.  
Ideally, $\ff$ should be chosen to maximally decrease the KL divergence with $p$; this can be framed as the following optimization problem: 
\vspace{-6pt}
\begin{align}\label{equ:diff}
\ff^* = \argmax_{\ff \in \F} 
\bigg\{\KL(q_0 ~||~p) - \KL(q_{[\epsilon \ff]} ~|| ~ p) \bigg \},
\end{align}
where $q_{[\epsilon \ff]}$ is the (empirical) distribution of $x^\prime=x+\epsilon \ff(x)$ when $x\sim q_0$, and $\F$ is a predefined function space that we optimize over.  
Note that although $\KL(q_0 ~||~p)$ can be infinite (or illy defined) when $q_0$ is an empirical delta measure, 
the difference of KL divergence in Eq. \eqref{equ:diff} can be finite because the infinite parts cancel out with each other. 
This can be checked by first approximating $q_0$ by a Gaussian mixture with variance $\sigma$, and 
then show that the limit exists and is finite when taking $\sigma$ to zero.

Equation \eqref{equ:diff} defines a challenging nonlinear functional optimization problem. 
It can be simplified by assuming the step size $\epsilon \rightarrow 0$,
in which case the decreasing rate of KL divergence 
can be approximated by the gradient of the KL divergence w.r.t. $\epsilon$ at $\epsilon =0$, that is, 
\begin{align}
	\ff^* = \argmax_{\ff \in\F} \bigg\{  -\frac{\dno}{\dno \epsilon} \KL(q_{[\epsilon \ff]} \parallel p) \big |_{\epsilon=0} \bigg \}.
	\label{equ:max-kl}
\end{align}
Further, \citet{liu2016stein} showed that the gradient objective in (\ref{equ:max-kl}) can be expressed as a linear functional of $\ff$, 
\begin{align*}
&-\frac{\dno}{\dno \epsilon}   \KL(q_{[\epsilon\ff]}~||~p)  \big |_{\epsilon =0} = \E_{x\sim q_0}[\stein_p \ff(x)] \\
&\text{with} ~~ \stein_p \ff(x) ~~\eqdef ~~\la \nabla_x \log p(x),~ \ff(x)\ra +\la \nabla_x, ~ \ff(x) \ra, 
\end{align*}
where $\stein_p$ is a linear operator acting on a $d\times 1$ vector-valued function $\ff$ and returns a scalar-valued function, and 
$\stein_p$ is called the \emph{Stein operator} in
connection with the so called Stein's identity, which says that $\E_{x\sim q} [\stein_p \ff(x)]=0$ when $q = p$ as a result of integration by parts.

Therefore, the optimization in (\ref{equ:max-kl}) reduces to
\begin{align}
\label{equ:ksd_prob}
\S(q_0 ~||~ p) ~~\eqdef ~~ \max_{\ff \in \F} \bigg\{ \E_{x\sim q_0} [\stein_p \ff(x)] \bigg\}, 
\end{align}
where $\S(q_0 ~||~ p)$ provides a notation of discrepancy measure between $q_0$ and $p$ and is known as the Stein discrepancy. 
If $\F$ is taken to be rich enough, $\S(q_0 ~||~p) = 0$ only if there exists no velocity field $\ff$ that can decrease the KL divergence between $p$ and $q_0$,   
which must imply $p=q_0$. 

The problem can be further simplified by taking a set $\F$ to have ``simple'' structures, but still remain to be infinite dimensional to catch all the possible useful velocity fields.  
A natural choice, motivated by kernel methods \citep[e.g.,][]{scholkopf2001learning}, is to take $\F$ to be the unit ball of a vector-valued reproducing kernel Hilbert space (RKHS) $\H = \H_0\times \cdots \times \H_0$, where each $\H_0$ is a scalar-valued RKHS associated with a positive definite kernel $k(x,x')$, that is, 
$$
\F = \{ \ff \in \H \colon ||\ff ||_{\H}  \leq 1\}. 
$$
Briefly speaking, $\H$ is the closure of functions of form $\ff(x) = \sum_{i}\vv a_i k(x,x_i)$, $\forall \vv a_i \in \RR^d$, $x_i \in \RR^d$, 
equipped with norm $||\ff ||_{\H}^2  = \sum_{ij}\vv a_i^\top \vv a_j k(x_i, x_j)$. 
With this choice, \citet{liu2016kernelized} showed the optimal solution of (\ref{equ:ksd_prob}) is $\ff^*/|| \ff^*||$, where 
\begin{align}
\ff^*(x^\prime) & = \E_{x\sim q_0} [\stein_p \otimes k(x,x^\prime)] \nonumber \\
              & = \E_{x\sim q_0}[ \nabla_x \log p(x) k(x,x^\prime) + \nabla_x k(x,x^\prime)].
\label{equ:phistar}
\end{align}
where $\stein_p \otimes f \overset{def}{=} \nabla \log p(x) f(x) + \nabla_x f(x)$ denotes the outer product version of Stein operator, 
which acts on a scalar-valued function $f$ and outputs a $d\times 1$ vector-valued function (velocity field).  
%

Therefore, $\ff^*$ provides the \emph{best} update direction within RKHS $\H$. 
By repeatedly applying this update starting with a set of initial particles,
 we obtain the SVGD algorithm: 
\begin{align}
\label{equ:svgd_update}
& x_i \gets x_i + \epsilon \ff^*(x_i), ~~~ ~~\text{} ~~~~ \forall i = 1,\ldots, n, \\ 
& \ff^*(x_i) =\frac{1}{n}\sum_{i=1}^n[\nabla_{x_j} \log p(x_j) k(x_j, x_i) + \nabla_{x_j} k(x_j,x_i)]. \nonumber
\end{align}
Update (\ref{equ:svgd_update}) mimics a gradient dynamics at the particle level, where the two terms in $\ff^* (x_i)$ play different roles: 
the term with the gradient $\nabla_x \log p(x)$ drives the particles toward the high probability regions of $p(x)$,
while the term with $\nabla_x k(x,x_i)$ serves as a repulsive force to encourage diversity as shown in \citet{liu2016stein}. 

It is easy to see from \eqref{equ:svgd_update} that $\ff^*(x_i)$ reduces to the typical gradient $\nabla_x \log p(x_i)$ when there is only a single particle ($n=1$) and $\nabla_x k(x,x_i)=0$ when $x=x_i$, 
in which cases SVGD reduces to the standard gradient ascent for maximizing $\log p(x)$ (i.e., maximum \emph{a posteriori} (MAP)).

\subsection{Learning Energy Models}

SVGD is an \emph{inference} process in which we want to find a set of particles (or ``data'') $\{x_i\}_{i=1}^n$ to approximate a given distribution $p(x)$. 
The goal of this paper is to investigate the \emph{learning} problem, the opposite of inference, in which we are given a set of observed data $\{x_i\}_{i=1}^n$
and we want to construct a distribution $p$, found in a predefined distribution family, to best approximate the data. 

In particular, we assume the observed data $\{x_i\}_{i=1}^n$ is i.i.d. drawn from an unknown distribution $p_\theta = p(x \mid \theta)$ indexed by a parameter $\theta$, of form  
\begin{align}
\begin{split}
& p(x\mid \theta) = \frac{1}{Z(\theta)}\exp(f(x; ~\theta)),  \\
& \text{with} ~~~Z(\theta) = \int_x \exp(f(x; ~\theta)) \dno x,
\end{split}
\label{eq:energy}
\end{align}
where $f(x; ~\theta)$ is a 
scalar-valued function that represents the negative energy of the distribution, 
and $Z(\theta)$ is the partition function which normalizes the distribution.  
%
A fundamental approach for estimating $\theta$ is the maximum likelihood estimation (MLE): 
\begin{align}
\hat{\theta} = \argmax_{\theta}\bigg \{ L(\theta ~|~ X) \equiv \frac{1}{n}\sum_{i=1}^n \log p(x_i\mid \theta) \bigg\}, 
\label{eq:mle}
\end{align}
where $L(\theta ~|~ X)$ is the log-likelihood function. 
For the energy-based model in \eqref{eq:energy}, 
the gradient of $L(\theta ~|~ X)$ can be shown to be 
\begin{align}\label{equ:grad}
\nabla_\theta {L(\theta \mid X)}=  \E_{q_0} [\nabla_\theta f(x; \theta)]  -  \E_{p_\theta} [\nabla_\theta f(x; \theta) ],
\end{align}
where we still use $q_0(x)=\frac{1}{n} \sum_{i=1}^n \delta(x - x_i)$ to denote the empirical distribution of data $\{x_i\}_{i=1}^n$. 
This gives a gradient ascent update for solving MLE: 
$$
\theta \gets \theta + \mu (\E_{q_0} [\nabla_\theta f(x; \theta)]  -  \E_{p_\theta} [\nabla_\theta f(x; \theta) ]), 
$$
where $\mu$ is the step size.
%
Intuitively, this update rule iteratively decreases the energy of the observed data (or the \emph{positive samples}), 
while increases the energy of the \emph{negative samples}, drawn from the hypothesized model $p_\theta$.  
When the algorithm converges, we should have 
$\nabla_\theta {L(\theta \mid X)} = 0$,
which is a moment matching condition between the empirical and the model-based averages of $\nabla_\theta f(x; \theta)$. 

However, critical computational challenges arise, 
because it is intractable to  exactly calculate the model-based expectation $ \E_{p_\theta} [\nabla_\theta f(x; \theta) ]$ and efficient approximation is needed. 
One way is to use Markov chain Monte Carlo  (MCMC) to approximate the expectation \citep[e.g.,][]{geyer1991markov,snijders2002markov}.
Unfortunately, MCMC-MLE is often too slow in practice, given that we need to approximate the expectation repeatedly at each gradient update step. 

\newcommand{\CD}{\mathrm{CD}}
\paragraph{Contrastive Divergence}
Maximum likelihood estimation can be viewed as 
finding the optimal $\theta$
to minimize the KL divergence between the empirical data distribution $q_0$ and the assumed model $p_\theta$: 
\begin{align}\label{equ:mleKL}
\min_{\theta} \KL(q_0 ~||~ p_\theta).  
\end{align}
Contrastive divergence \citep{hinton2002training} is an alternative method that optimizes a different objective function:
\begin{align}\label{equ:cdk}
\min_\theta \big\{
\CD_k \overset{def}{=}
\KL(q_0 ~||~ p_\theta) - \KL(q_k ~|| ~ p_\theta) \big\}, 
\end{align}
where $q_k$ is a distribution obtained by 
\emph{moving $q_0$ towards $p_\theta$ for k steps}, more precisely, by running $k$ steps of Markov transitions whose equilibrium distribution $p_\theta$, starting from the empirical distribution $q_0$. 
$\CD_k$ is always non-negative because running Markov chain forward can only decrease the KL divergence \citep{cover2012elements}, 
and equals zero only if $q_0$ matches $p_\theta$. 
Observe that Eq. \eqref{equ:cdk} and \eqref{equ:diff} share a similar objective function, but optimize different variables ($\ff$ vs. $\theta$). 
Their similarity is the main motivation of our Stein contrastive divergence algorithm (Section~\ref{sec:steincd}). 

Taking gradient descent on the $\CD_k$ objective (without differentiating through $q_k$) gives the following update rule: 
\begin{align}\label{equ:cdgrad}
\theta \gets \theta + \mu (  \E_{q_0} [\nabla_\theta f(x; \theta)  ] -  \E_{q_k} [ \nabla_\theta f(x; \theta)]  ). 
\end{align}
Compared with the MLE update \eqref{equ:grad}, 
the CD-$k$  update replaces the ``ideal'' negative sample drawn from $p_\theta$ with a local 
$k$-step perturbation of the observed data. 
Although CD-$\infty$ can be viewed as MLE, 
the key observation of \citet{hinton2002training} is that even by using a small $k$, such as $k = 1$, we obtain useful contrastive information about how the parameter $\theta$ should be improved. 

\section{Stein Contrastive Divergence}
\label{sec:steincd}

\begin{algorithm}[t]
\caption{Stein Contrastive Divergence (SteinCD)}
\label{alg:steincd}
\begin{algorithmic}
\STATE {\bf Goal:} Learn energy model \eqref{eq:energy}  from data $\{x_i\}_{i=1}^n$. 
\WHILE{no Converged} 
\STATE 1.~Draw a minibatch of positive sample $\{x_i^+\}_{i=1}^m$ from the training set. 
\STATE 2.~Perform one step of SVGD update (Eq. \ref{equ:svgd_update}) on $\{x_i^+\}_{i=1}^m$ 
to get negative samples $x_i^-$ by 
$$x_i^- \leftarrow x_i^+ + \epsilon \ff^*(x_i^+),~~~\forall i = 1,\ldots, m.$$
\STATE 3.~Update $\theta$ by 
$$
\theta \gets  \theta  + \frac{\mu}{m}\sum_{i=1}^m( \nabla_\theta f(x_i^+;\theta) - \nabla_\theta f(x_i^-;\theta)). 
$$
\ENDWHILE
\end{algorithmic}
\end{algorithm}

The performance of CD depends on the choice of Markov chain it uses; 
it is clear that we should select the Markov chain 
to minimize $\KL(q_k ~||~ p_\theta)$ and hence maximize the contrastive objective \eqref{equ:cdk}, bringing it closer to the MLE objective \eqref{equ:mleKL}.  
However, it is unclear how to frame the optimal choice of Markov chains into an solvable optimization problem. 
SVGD provides a natural solution for this, given that it explicitly provides the best perturbation direction that maximizes the very same contrastive objective.  

To be more specific, 
assume we perturb the observed data $\{x_i\}_{i=1}^n$ with a deterministic transform $ x^\prime \gets x + \epsilon \ff(x)$ as in SVGD, 
where the velocity field $\ff$ is decided jointly with the model parameter $\theta$ by solving the following minimax problem: 
\begin{align}\label{equ:minmax}
\min_{\theta} \max_{\ff \in \F } \big \{\frac{1}{\epsilon} (\KL(q_0 ~||~ p_\theta) - \KL(q_{[\epsilon \ff]} ~||~ p_\theta)) \big \}. 
\end{align}
We then solve this problem by alternating between
minimizing $\theta$ and maximizing $\ff$. 
Following the derivation of SVGD,  
with small step size $\epsilon$, 
$\ff$ has a closed form solution shown in \eqref{equ:phistar}, 
and the gradient update of $\theta$ is 
\begin{align}\label{equ:steincd}
\theta \gets \theta + \mu(\E_{q_0}[\nabla_\theta f(x ; \theta)] - \E_{q_{[\epsilon \ff^*]}}[\nabla_\theta f(x ; \theta)]), 
\end{align}
where we update $\theta$ using the result of one step of SVGD update on the observed data as the negative samples. 
This gives our \emph{Stein contrastive divergence} shown in Algorithm~\ref{alg:steincd}, 
which replaces the $k$-step Markov chain perturbation in typical CD with a SVGD update.

\paragraph{Stein Score Matching}
We should keep the step size $\epsilon$ small to make the derivation valid.
If we explicitly take $\epsilon \to 0$, 
then the minimax problem in \eqref{equ:minmax} reduces to minimizing the Stein discrepancy between the data distribution 
$q_0$ and model 
$p_\theta$, that is, 
\begin{align}\label{equ:mins}
\min_\theta \S^2(q_0 ~||~ p_\theta),  
\end{align}
which is the result of Eq.~\eqref{equ:max-kl} and \eqref{equ:ksd_prob}. 
From this perspective, 
it is possible to take \eqref{equ:mins} and directly derive a gradient descent algorithm for minimizing the Stein discrepancy
\eqref{equ:mins}.
From $\S^2(q_0 ~||~p) = \E_{q_0}[\stein_p\ff^*]$,  we can derive that  
\begin{align*}
\nabla_{\theta} \S^2(q_0 ~||~ p_\theta) 
=  2 \E_{x\sim q_0}[\nabla_\theta \nabla_x f (x; \theta) \ff^*(x)], 
\end{align*}
where $\nabla_\theta \nabla_x  f(x; ~ \theta)$ is the $(m \times d)$-valued cross derivative of $\log p(x|\theta),$ where $m$ and $d$ are the dimensions of $\theta$ and $x$, respectively. 
This gives the following update:   
\begin{align}\label{equ:steinsm}
\theta \gets  \theta - \mu  \E_{q_0}[\nabla_\theta \nabla_x f (x; \theta) \ff^*(x)]. 
\end{align}
We call this update rule \emph{Stein score matching} in connection with the score matching algorithm \citep{hyvarinen2005estimation} which minimizes Fisher divergence. 

In practice, it can be cumbersome to  calculate the cross derivative $\nabla_\theta \nabla_x  f(x; \theta)$. 
It turns out the SteinCD update \eqref{equ:steincd} can be viewed as approximating $\nabla_\theta \nabla_x  f(x;\theta)$ in \eqref{equ:steinsm} with a finite difference approximation: 
\begin{align}
& \nabla_\theta \nabla_x f(x; \theta) \ff^*(x)   \notag \\
&~~ \approx  - \frac{1}{\epsilon}\big[ \nabla_\theta  f(x; ~\theta)  - \nabla_\theta f (x + \epsilon \ff^*(x); ~\theta)) \big]. 
\label{equ:adfd}
\end{align}
Plugging the above approximation into \eqref{equ:steinsm} gives \eqref{equ:steincd}. 

Alternatively, it is also possible to use a symmetric finite difference formula:
\begin{align*}
& \nabla_\theta \nabla_x  f(x; \theta) \ff^*(x)   \\
&~~ \approx  -\frac{1}{2\epsilon} ((\nabla_\theta  f(x - \epsilon \ff^*(x); ~\theta)  - \nabla_\theta f (x + \epsilon \ff^*(x); ~\theta)). 
\end{align*}
This corresponds to 
\begin{align}\label{equ:steincd_adv}
\theta \gets \theta + \mu(\E_{q_{[-\epsilon\ff^*]}}[\nabla_\theta f(x ; \theta)] - \E_{q_{[\epsilon \ff^*]}}[\nabla_\theta f(x ; \theta)]),
\end{align}
which perturbs the data on both opposite directions and uses the difference to guild the update of $\theta$. 

In practice, we prefer the original update \eqref{equ:steincd} 
because of its simplicity. 
Note that unlike SVGD, there is no cost in using a small step size beyond the error caused by numerical rounding, 
since we just need to obtain a correct moving direction, do not need to actually move the particles (the observed data) to $p_\theta$. 
Therefore, we can use a relatively small $\epsilon$,  in which case (\ref{equ:steincd}), (\ref{equ:steinsm}) and (\ref{equ:steincd_adv}) are all close to each other. 

It is worth comparing Stein score matching \eqref{equ:steinsm} 
with the (Fisher) score matching \citep{hyvarinen2005estimation}, which estimates $\theta$ by minimizing the Fisher divergence:
\begin{align}\label{equ:fisher}
\!\!\!\min_\theta\big\{ \mathbb F(q_0 ~||~ p_\theta) \overset{def}{=} \E_{q_0}[||\nabla_x \log q_0 - \nabla_x \log p_\theta||^2_2]\big\}. 
\end{align}
Fisher divergence is a stronger divergence measure than Stein discrepancy, which 
equals infinite (like KL divergence) if $q_0$ is an empirical delta measure, because $\nabla_x\log q_0$ does not exist for delta measures. 
In contrast, Stein discrepancy remains to be finite for empirical measure $q_0$ because it depends on $q_0$ only through the empirical averaging $\E_{q_0}[\cdot]$. 

Nevertheless, like the case of KL divergence, 
the infinite part of Fisher divergence does not depend on $\theta$, and it is still possible to minimize \eqref{equ:fisher} as shown in \citet{hyvarinen2005estimation}, by using integration by parts. 
The main disadvantage of Fisher score matching is that it has a relatively complex formula, 
and involves calculating a third order derivative ${\partial^3\log p(x|\theta)}/{\partial\theta \partial x^2}$ that makes it difficult to implement. 
In contrast, SteinCD only involves calculating the first order derivatives and is straightforward to implement. 

%


\section{Amortized MLE }
\label{sec:steingan}
CD-type algorithms 
have been widely used 
for learning energy-based models, 
and can often train models with good test likelihood. 
However, models trained by CD can not generate realistic looking images (when $p_\theta$ is used to model image pixels).
This is because CD learns the models based on a local perturbation 
in the neighborhood of the observed data, 
and does not explicitly train the model to create images from scratch. 
This problem has been addressed recently 
by generative adversarial networks (GAN) \citep{goodfellow2014generative,radford2015unsupervised}, 
which explicitly train a generator, a deep neural network that takes random noise and outputs
images, to match the observed data with the help of a discriminator
that acts adversarially, to distinguish the generated data from the observed ones. 
%
Motivated by GAN, 
we modify the MLE and CD idea to explicitly incorporate a generator into the training process. 

Our idea is based on ``amortizing'' the sampling process of $p_\theta$ with a generator 
and use the simulated samples as the negative samples to update $\theta$. 
To be specific, let $G(\xi; \eta)$ be a neural network that takes a random noise $\xi$ as input and outputs a sample $x$, 
with a parameter $\eta$ which we shall adjust adaptively to make the distribution of $x = G(\xi; ~\eta)$ approximates the model $p_\theta  = p(x|\theta)$,
and we update $\theta$ by 
\begin{align}\label{equ:thetaG}
\theta \gets \theta + \mu (\E_{p_0}[\nabla f(x ; \theta)] - \E_{G_\eta}[\nabla f(x; \theta)]), 
\end{align}
where $\E_{G_\eta}[\cdot ]$  denotes the average on random variable $x = G(\xi; ~\eta).$ 
The key question here is how to 
update $\eta$ so that the distribution of $x = G(\xi; ~\eta)$ closely approximates the target distribution $p_\theta$. 
This problem is addressed by a recent amortized SVGD algorithm \citep{wang2016steingan}, as we introduce as follows. 
\paragraph{Amortized SVGD}
The idea of amortized SVGD is to leverage the Stein variational gradient direction \eqref{equ:max-kl} to guid the 
update of the generator $G(\xi; ~\eta)$, in order to match its output distribution with $p_\theta$.  
To be specific, at each iteration of amortized SVGD, we generate 
a batch of random outputs $\{x_i\}$ where $x_i = G(\xi_i; ~\eta)$, based on the current parameter $\eta$. 
The Stein variational gradient $\ff^*(x_i)$ in \eqref{equ:svgd_update}
 would then ensure that $x_i' = x_i + \epsilon\ff^*(x_i)$ 
 forms a better approximation of the target distribution $p_\theta$. 
 Therefore, we should adjust $\eta$ to make the output of the generator match the updated points ${x_i'}$. 
This can be done by updating $\eta$ via 
\begin{align}\label{equ:amort0}
\eta \gets \argmin_{\eta} \sum_{i=1}^n ||  G(\xi_i;~\eta) - x_i - \epsilon \ff^*( x_i) ||_2^2.
\end{align}
Essentially, this projects the non-parametric perturbation direction $\ff^*(x_i)$ to the change of the finite dimensional network parameter $\eta$. 
Assume the step size $\epsilon$ is taken to be small, 
so that a single step of gradient descent provides a good approximation of Eq. \eqref{equ:amort0}. 
This gives a simpler update rule: 
\begin{align}\label{equ:amort1}
\eta \gets  \eta + \epsilon \sum_i \partial_\eta G(\xi_i; \eta) \ff^*(x_i),
\end{align}
which can be intuitively interpreted as a form of chain rule that \emph{back-propagates the SVGD gradient to the network parameter $\eta$}.  
In fact, when there is only one particle, \eqref{equ:amort1} 
reduces to the standard gradient ascent for $\max_{\eta} \log p_\theta(G(\xi;~\eta))$,
in which $G_\eta$ is trained to ``learn to optimize'' \citep[e.g.,][]{andrychowicz2016learning}, instead of ``learn to sample'' $p_\theta$. 
Importantly, as there are more than one particles, 
the repulsive term $\nabla_x k(x,x_i)$ in $\ff^*(x_i)$  becomes active, and enforces an amount of diversity on the network output that is consistent with the variation in $p_\theta$. 


\paragraph{Amortized MLE}
Updating $\theta$ and $\eta$ alternatively with \eqref{equ:thetaG} and \eqref{equ:amort0} or \eqref{equ:amort1} 
allows us to simultaneously train an energy model together with a generator (sampler). 
This approach is presented as Algorithm \ref{alg:steingan}.
Compared with the traditional methods based on MCMC-MLE or contrastive divergence, 
we \emph{amortize the sampler as we train}, which saves computation in long term 
and simultaneously provides a high quality generative neural network that can generate realistic-looking images. 

Formally, our method can be viewed as approximately solving the following minimax objective function based on KL divergence:   
\begin{align}\label{equ:gam}
\min_\theta \max_{\eta} \big\{ \KL(q_0 ~||~ p_\theta) - \KL(q_{[G_\eta]} ~||~ p_\theta) \big \},
\end{align}
where $q_{[G_\eta]}$ denotes the distribution of the output of the generator $x = G(\xi; ~\eta)$.  
Here the energy model $p_\theta$, 
serving as a discriminator,  attempts to get closer to the observed data $q_0$, and keep away from the ``fake'' data distribution $q_{[G_\eta]}$, both in terms of 
KL divergence,
 while the generator $G_\eta$ attempts to get closer to the energy model $p_\theta$ using amortized SVGD. 
 We call our method \emph{SteinGAN} to reflect this interpretation. 
 It can be viewed as a KL-divergence variant of the GAN-style adversarial game \citep{goodfellow2014generative}. 



\begin{algorithm}[t]
\caption{Amortized MLE (also called SteinGAN)}
\label{alg:steingan}
\begin{algorithmic}   
\STATE {\bf Goal:} Learn energy model \eqref{eq:energy}  from data $\{x_i\}_{i=1}^n$. 
\WHILE{not converged}
\STATE 1. Draw minibatch $\{ x_i^+\}_{i=1}^m$ from the training data. 
\STATE 2. Draw $\{\xi_i\}_{i=1}^m$ from the noise prior. Calculate the negative sample $x_i^- = G(\xi_i; \eta)$, $i=1,\ldots,m$.
\STATE 3. Update the generator parameter $\eta$ by 
$$
\eta \gets \eta+ \frac{\epsilon}{m} \sum_{i=1}^m [\partial_\eta G(\xi_i; \eta) \ff^{*} (x_i^-) ],~~~~~~\text{where}
$$
$$
\ff^{*}(\cdot)
= \frac{1}{m}\sum_{j=1}^m 
[  \nabla_{x_j^-} f(x_j^-;\theta) k(x_j^-, \cdot) + \nabla_{x_j^-}k(x_j^-, \cdot)]. 
$$
\STATE 4. Update the parameter $\theta$ 
\begin{align}\label{steindd}
	\theta \gets \theta  + \frac{\mu}{m}\sum_{i=1}^m(\nabla_\theta f(x_i^+; \theta) -  \nabla_\theta f(x_i^-; \theta)).
\end{align}
\ENDWHILE
\end{algorithmic}   
\end{algorithm}

%
%


\paragraph{Related Work}
The idea of training energy models with neural samplers was also discussed by \citet{kim2016deep, zhai2016generative}; 
one of the key differences is that the neural samplers in \citet{kim2016deep, zhai2016generative} 
are trained with the help of heuristic diversity regularizers, 
 while SVGD enforces the diversity in a more principled way. 
Another method by \citet{zhao2016energy} also trains an energy function to distinguish real and simulated samples, but within a non-probabilistic framework. 

Generative adversarial network (GAN) and its variants have recently gained remarkable success in generating realistic-looking images 
\citep[][to name a few]{goodfellow2014generative, salimans2016improved, radford2015unsupervised, li2015generative,dziugaite2015training,nowozin2016f, wgan}. 
All these methods are set up to train implicit models specified by the generators, 
and are different from the energy model assumption. 
The main motivation of SteinGAN is to show the possibility of obtaining comparable image generation results 
using energy-based learning, 
allowing us to combine the advantages of these two types of approaches. 


\begin{figure*}
\centering
\begin{tabular}{cccc}
\raisebox{1.3em}{\rotatebox{90}{\small test data log likelihood}} \includegraphics[height=0.2\textwidth]{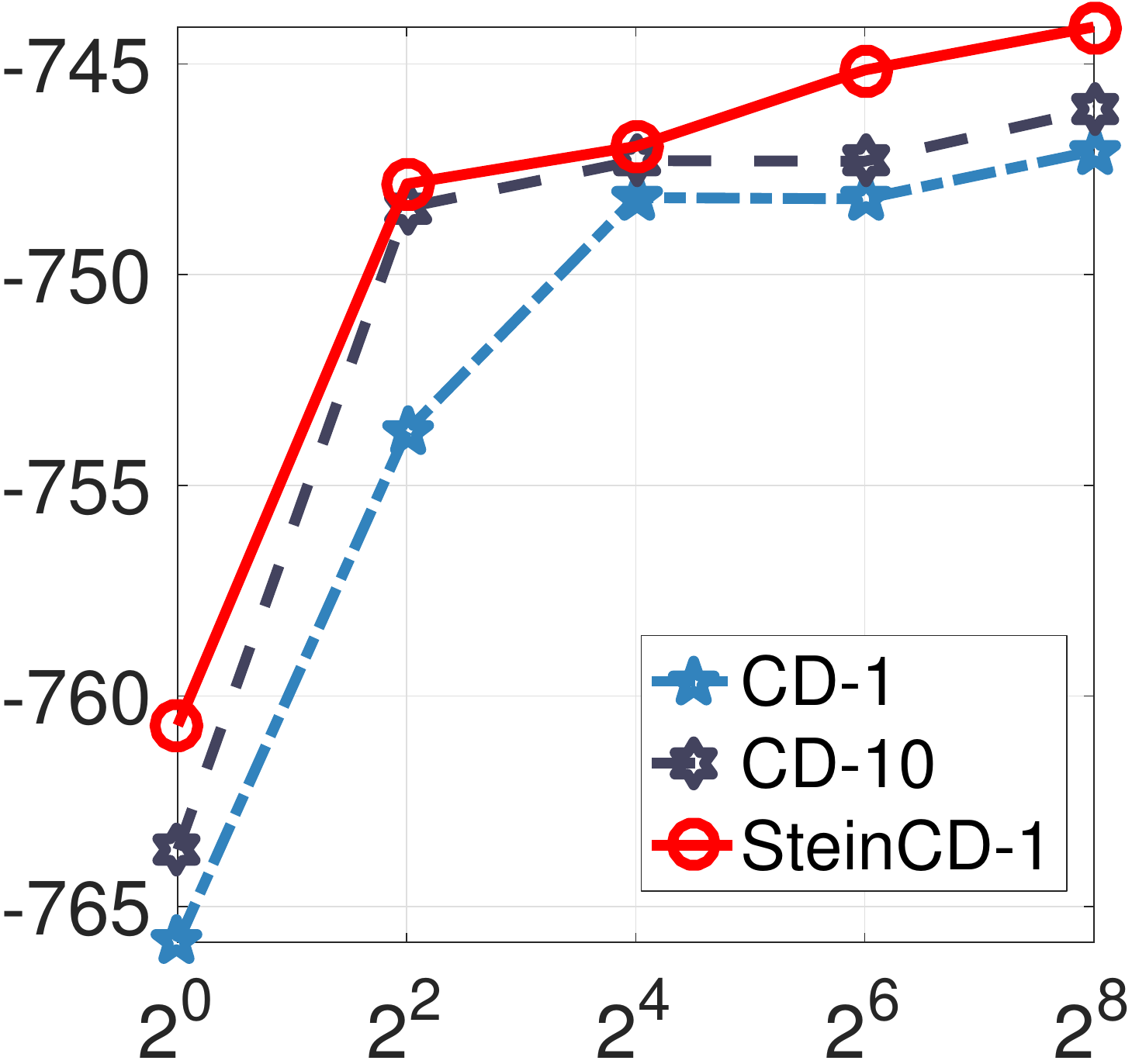} &
\raisebox{2.3em}{\rotatebox{90}{\small inception score}} \includegraphics[height=0.2\textwidth]{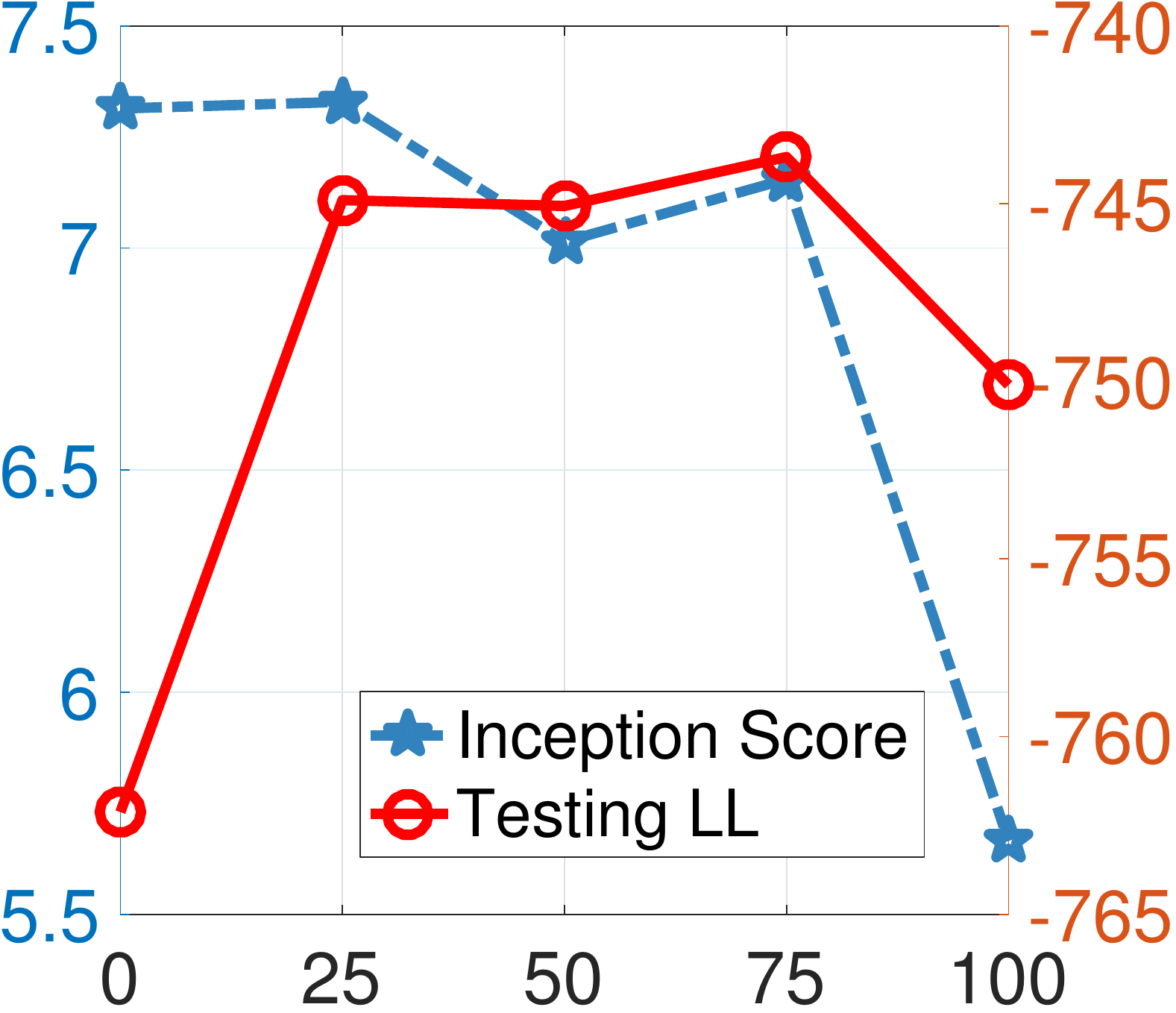}
\raisebox{1.3em}{\rotatebox{90}{\small test data log likelihood}} &
\hspace{.5em}
\includegraphics[height=0.2\textwidth]{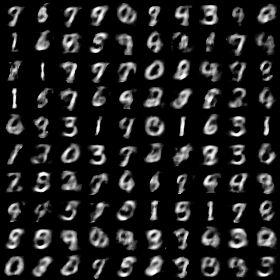} &
\includegraphics[height=0.2\textwidth]{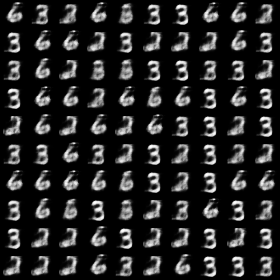} \\
(a)\small ~~ Iterations ($\times10$) &  (b) \small ~~CD update ratio ($100\alpha\%$) &(c) ~~\small SteinGAN & (d)~\small SteinGAN w/o kernel \\
\end{tabular}
\vspace{-.8\baselineskip}
\caption{\small Results of RBM on MNIST. (a) test data log likelihood of SteinCD and typical CD-1 and CD-10 with Langevin dynamics. 
(b) Inception score and test data log likelihood of SteinGAN-CD($\alpha$) with different $\alpha$, which reduces to SteinGAN when $\alpha=0$ and SteinCD when $\alpha = 0$.   
(c) Samples generated by SteinGAN; 
(d) Samples generated by SteinGAN when the kernel is turned off.}
\label{fig:rbm}
\end{figure*}

\section{Experiments}
\label{sec:experiments}

We evaluated SteinCD and SteinGAN on four datasets,  
including MNIST, CIFAR-10, CelebA \citep{liu2015faceattributes}, and Large-scale Scene Understanding (LSUN) \citep{yu2015lsun}. 
We observe:  1) SteinCD tends to outperform typical CD equipped with Langevin dynamics; 
2) SteinCD tends to provide better test likelihood than SteinGAN, while SteinGAN generates better images; 
3) by interleaving SteinCD and SteinGAN updates, it is possible to combine advantages of both, obtaining both good testing likelihood and images. 
We will provide code to reproduce our experiments. 

\subsection{Restricted Boltzmann Machine on MNIST}
In this section, we evaluate our methods on MNIST and use a simple Gaussian-Bernoulli Restricted Boltzmann Machines (RBM) as our energy-based model, 
which allows us to accurately evaluate the test likelihood. 
Gaussian-Bernoulli RBM is a hidden variable model of form 
$$
p(x, h) = \frac{1}{Z} \exp(\frac{1}{2} x^\top Bh + b^\top x + c^\top h -\frac{1}{2} || x ||_2^2),
$$
where $x\in \R^d$ is a continuous observed variable and $h\in\{\pm 1\}^\ell$ is a binary hidden variable; 
$Z$ is the normalization constant. By marginalizing out the hidden variable $h$, we obtain  
$p(x) = \exp(f(x))/Z$ with negative energy 
$$
f(x) = b^\top x - \frac{1}{2}|| x ||^2 
+\sigma(B^\top x + c), 
$$
where $\sigma(h) = \sum_{i=1}^\ell\log(\exp(h_i) + \exp(-h_i))$.
To evaluate the test likelihood exactly, we use a small model with only $\ell=10$ hidden units which allows us to calculate $\log Z$
using exact variable elimination.  
We use mini-batches of size 100  and  Adam  \citep{kingma2014adam} for our gradient updates. 
Following \citet{liu2016stein}, we use a RBF kernel $k(x,x') = \exp(-||x-x'||^2_2/h)$ in SVGD, with the bandwidth $h$ selected to be $med^2 /\log m$ ($med$ is the median of the pairwise distance of $\{x_i\}$ and $m$ is the minibatch size). 
For SteinGAN, the generator consists of 2 fully connected layers, followed by 2 deconvolution layers with $5\times 5$ filters.

Figure~\ref{fig:rbm}(a) shows the test likelihood of the CD-type methods, 
where we can find that SteinCD (with a single step of SVGD) outperforms both CD-1 and CD-10 equipped with Langevin dynamics. 
We also evaluated the test likelihood of SteinGAN but find it gives worse test likelihood (not shown in the figure). 
The advantage of SteinGAN, however, 
is that it trains a generator $G(\xi; ~\eta)$ that produces high quality and diverse images (Figure~\ref{fig:rbm}(c)), 
which SteinCD can not generate (even when we train another generator $G(\xi; ~\eta)$ on the energy model obtained by SteinCD, which has better test likelihood).

\paragraph{Mixing SteinCD and SteinGAN}
It seems that SteinCD and SteinGAN have opposite properties in terms of test likelihood and image quality.
This motivates us to integrate SteinCD and SteinGAN to combine their advantages. 
In order to do this, we replace the neural-simulated sample $x =G(\xi; \eta)$ (step 4 in Algorithm \ref{alg:steingan})
with the SVGD-updated sample in SteinCD (step 2 of Algorithm~\ref{alg:steincd}) with a probability $\alpha$. 
We  use SteinCD-GAN($\alpha$) to denote this algorithm,  
which reduces to SteinCD with $\alpha = 1$ and SteinGAN with $\alpha=0$.

The performance of SteinCD-GAN($\alpha$)
with different $\alpha$-values is shown in Figure~\ref{fig:rbm}(b), where we find that mixing even a small percentage of CD-updates (e.g., $\leq 25\%$) 
can significantly improve the test likelihood, even slightly higher than the pure SteinCD algorithm. 

We also evaluate the inception score of the images generated by SteinCD-GAN($\alpha$), using an inception model trained on the MNIST training set (result averaged on 50,000 generated images). As shown in Figure~\ref{fig:rbm}(b), 
we find that adding CD updates seems to deteriorate the image quality, 
but only slightly unless we use $100\%$ CD updates ($\alpha = 1$). 
Overall, Figure~\ref{fig:rbm}(b) suggests that by mixing SteinGAN with a small percentage of CD updates (e.g., $\alpha = 25\%$), we obtain results that perform well both in terms of test likelihood and image quality. 

\paragraph{Effect of the Repulsive Force}
As it is discussed in Section~\ref{sec:svgd}, the repulsive term  $\nabla_x k(x,x')$ in SVGD \eqref{equ:svgd_update} 
enforces the particles to be different from each other and produces an amount of variability required for generating samples from $p(x)$.  
%
In order to investigate the effect of the repulsive term $\nabla_{x}k(x,x')$. 
We test a variant of SteinGAN in which the Stein variational gradient $\ff^*(x)$ is replaced by the typical gradient $\nabla_x \log p(x|\theta)$ (or, effectively, use a constant kernel $k(x,x') = 1$ in $\ff^*(x$));  
this corresponds to an amortized variant of Viterbi learning \citep{koller2009probabilistic}, or Herding algorithm \citep{welling2009herding} that maximizes 
$\E_{q_0} [\log p(x|\theta)]  - \max_{\eta} \E_{\xi}[\log p(G(\xi; \eta) | \theta)],$ as the learning objective function. 

As shown in Figure~\ref{fig:rbm}(d), SteinGAN without the kernel 
tends to produce much less diverse images. 
This suggests that the repulsive term is responsible for generating diverse images in SteinGAN. 

\newcommand{\fttmp}{\small}
\begin{figure*}[t]
\centering
\begin{tabular}{cccc}
\raisebox{6em}{
\renewcommand{\arraystretch}{1.07}
\newcommand{\tmpfnt}{\small}
\hspace{-.08\textwidth}
\begin{tabular}{r}
\tmpfnt airplane \\
\tmpfnt automobile \\
\tmpfnt bird \\
\tmpfnt cat \\
\tmpfnt deer \\
\tmpfnt dog \\
\tmpfnt frog \\
\tmpfnt horse \\
\tmpfnt ship \\
\tmpfnt truck
\end{tabular}} & 
\hspace{-.025\textwidth}
\includegraphics[width=.25\textwidth]{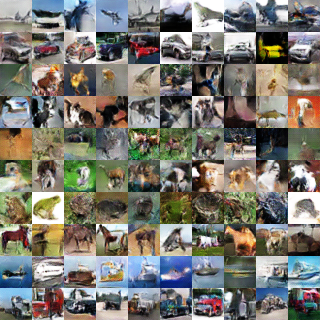}  & 
\includegraphics[width=.25\textwidth]{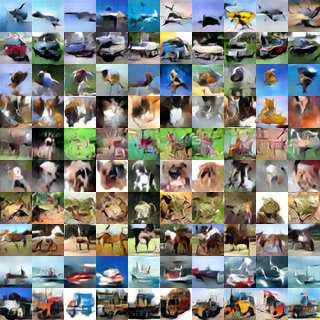}  &
\includegraphics[width=.25\textwidth]{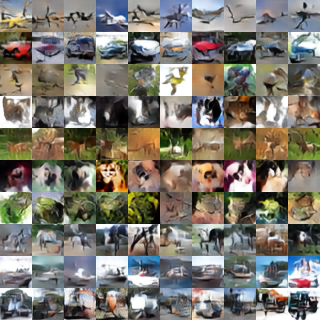}   \\
& \small DCGAN 
& \small SteinGAN & \small SteinGAN w/o kernel\\
\vspace{5\baselineskip}
\end{tabular}
\renewcommand{\arraystretch}{1.2}
\begin{tabular}{|l|c|c|c|c|c|}
\hline
 & \fttmp Real Training Set &  \fttmp  500 Duplicate &  \fttmp DCGAN  &  \fttmp SteinGAN  &  \fttmp SteinGAN w/o kernel\\
\hline
 \fttmp  Inception Score &  \fttmp 11.237 &  \fttmp 11.100 &  \fttmp 6.581 & \fttmp  6.711 & \fttmp   \fttmp 6.243 \\ 
\hline
 \fttmp   Testing Accuracy &  \fttmp 92.58 \% &  \fttmp 44.96 \% &   \fttmp 44.78 \% & \fttmp  61.09 \% &  \fttmp 60.50 \% \\
\hline
\end{tabular}  \\
\caption{
\small Results on CIFAR-10. ``500 Duplicate'' denotes 500 images randomly subsampled from the 
training set, each duplicated 100 times.
Upper: images simulated by DCGAN and SteinGAN (based on joint model \eqref{equ:pxy}) conditional on each category;
 Lower: Inception scores for samples generated by various methods (all with 50,000 images) on inception models trained on ImageNet \citep{salimans2016improved}, 
 and testing accuracies on the real testing set when 
train ResNet classifiers \citep{he2016deep} on 1) Real Training Set,  2) 100 copies of 500 examples taken at random from the real training set, 3) 50,000 samples from DCGAN, 4) 50,000 samples from SteinGAN, and 5) 50,000 samples from SteingGAN w/o kernel, respectively. 
We set $m=1$ in Eq.(\ref{equ:pxy}).
}
\label{fig:cifar10}
\end{figure*}

\begin{figure*}[t]
\centering
\begin{tabular}{cc}
\includegraphics[width=0.3\textwidth]{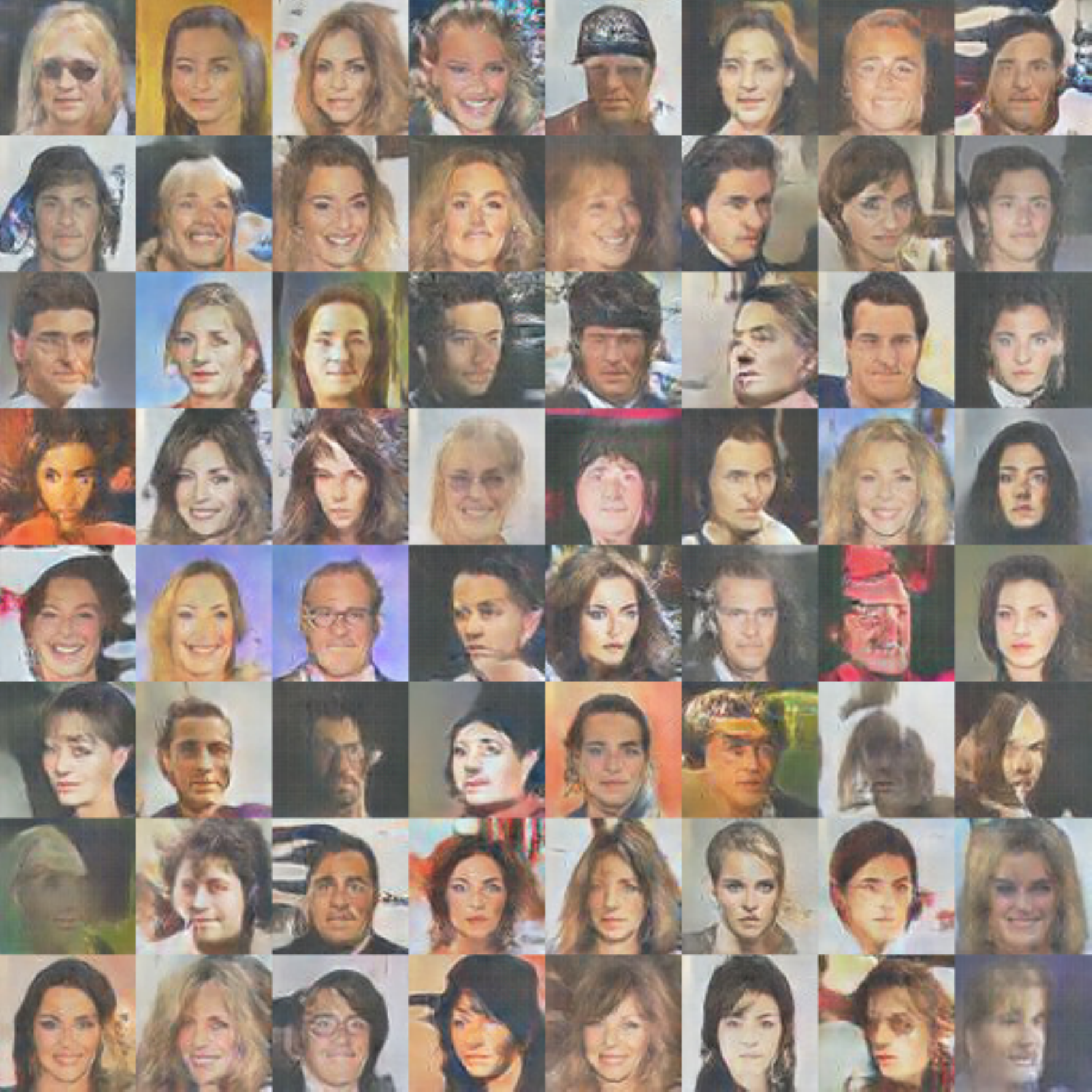} & 
\includegraphics[width=0.3\textwidth]{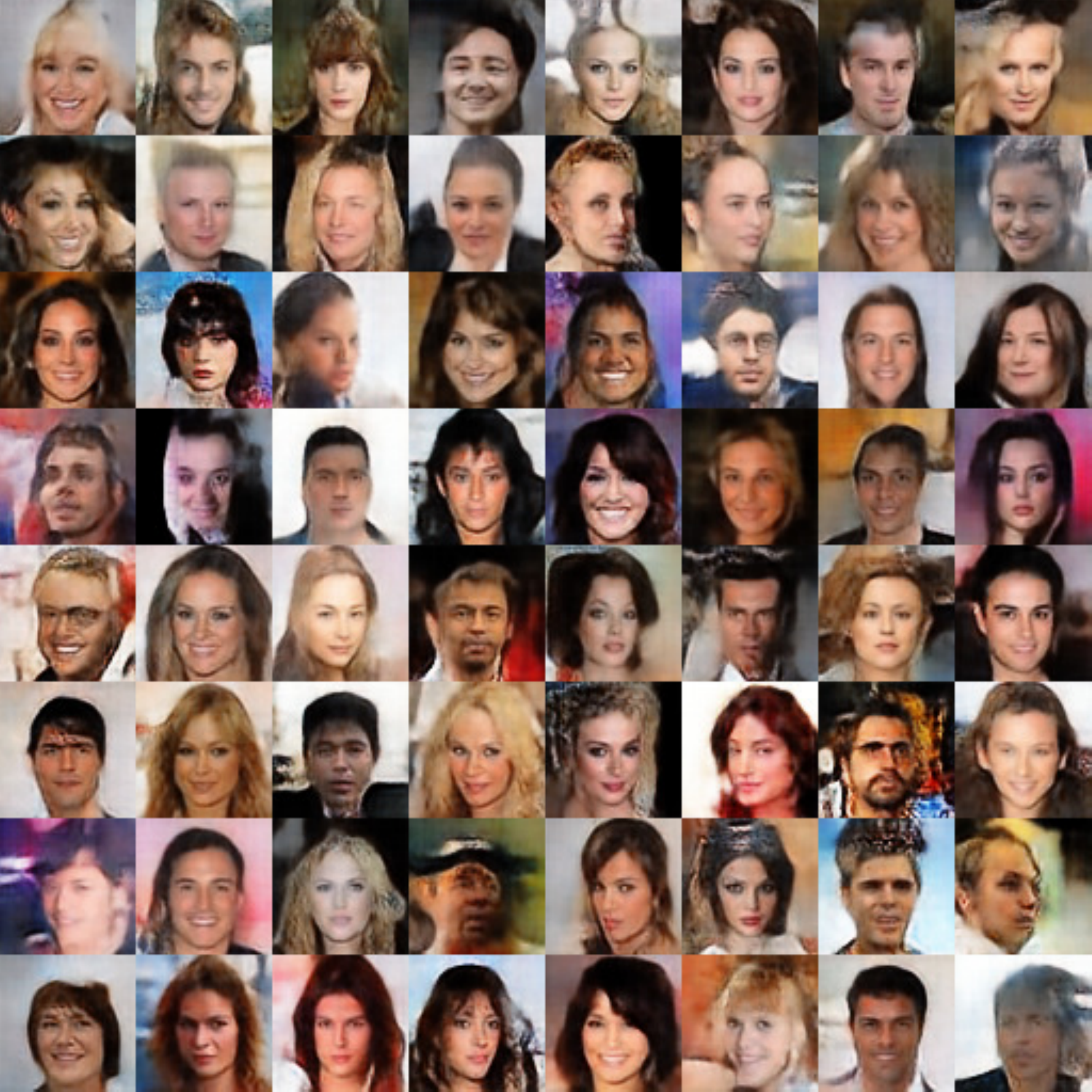} \\
\fttmp DCGAN & \fttmp  SteinGAN \\
\end{tabular}
\begin{tabular}{c}
\includegraphics[width=0.6\textwidth]{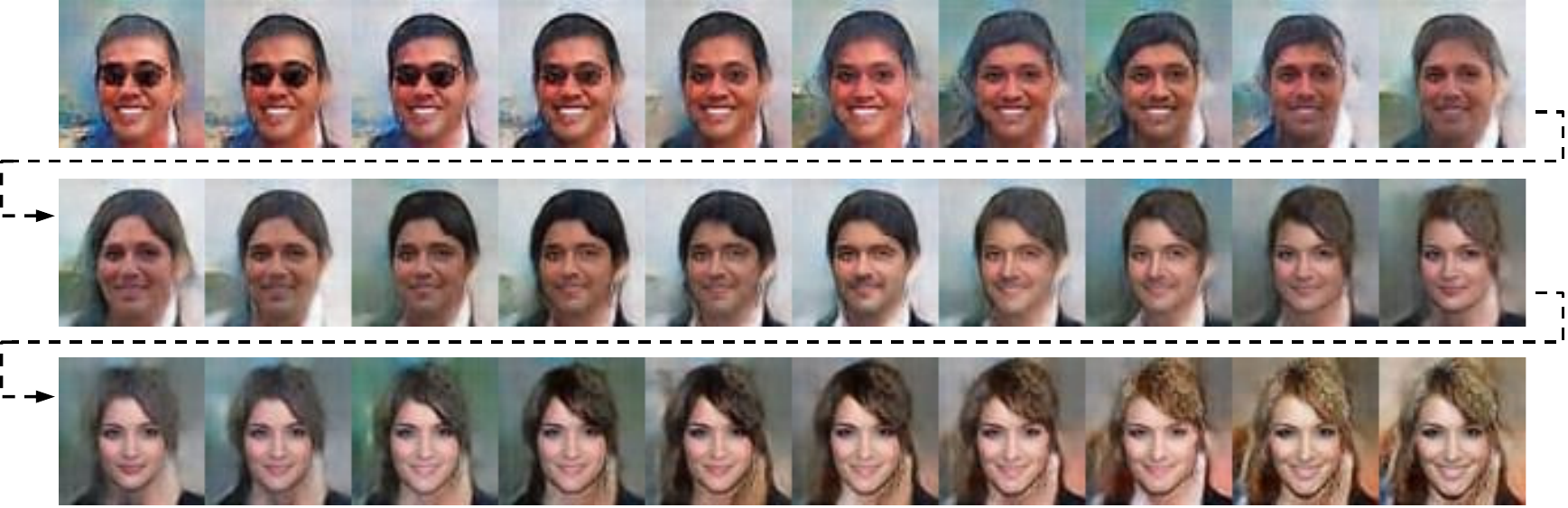} \\
\end{tabular}
\caption{\fttmp  Results on CelebA. Upper: images generated by DCGAN and our SteinGAN. Lower: images generated by SteinGAN when performing a random walk $\xi\gets \xi + 0.01\times\mathrm{Uniform}([-1,1])$ on the random input $\xi$; we can see that a man with glasses and black hair gradually changes to a woman with blonde hair. 
}
\label{fig:face}
\end{figure*}

\begin{figure*}[t]
\centering
\begin{tabular}{cc}
\includegraphics[width=0.3\textwidth]{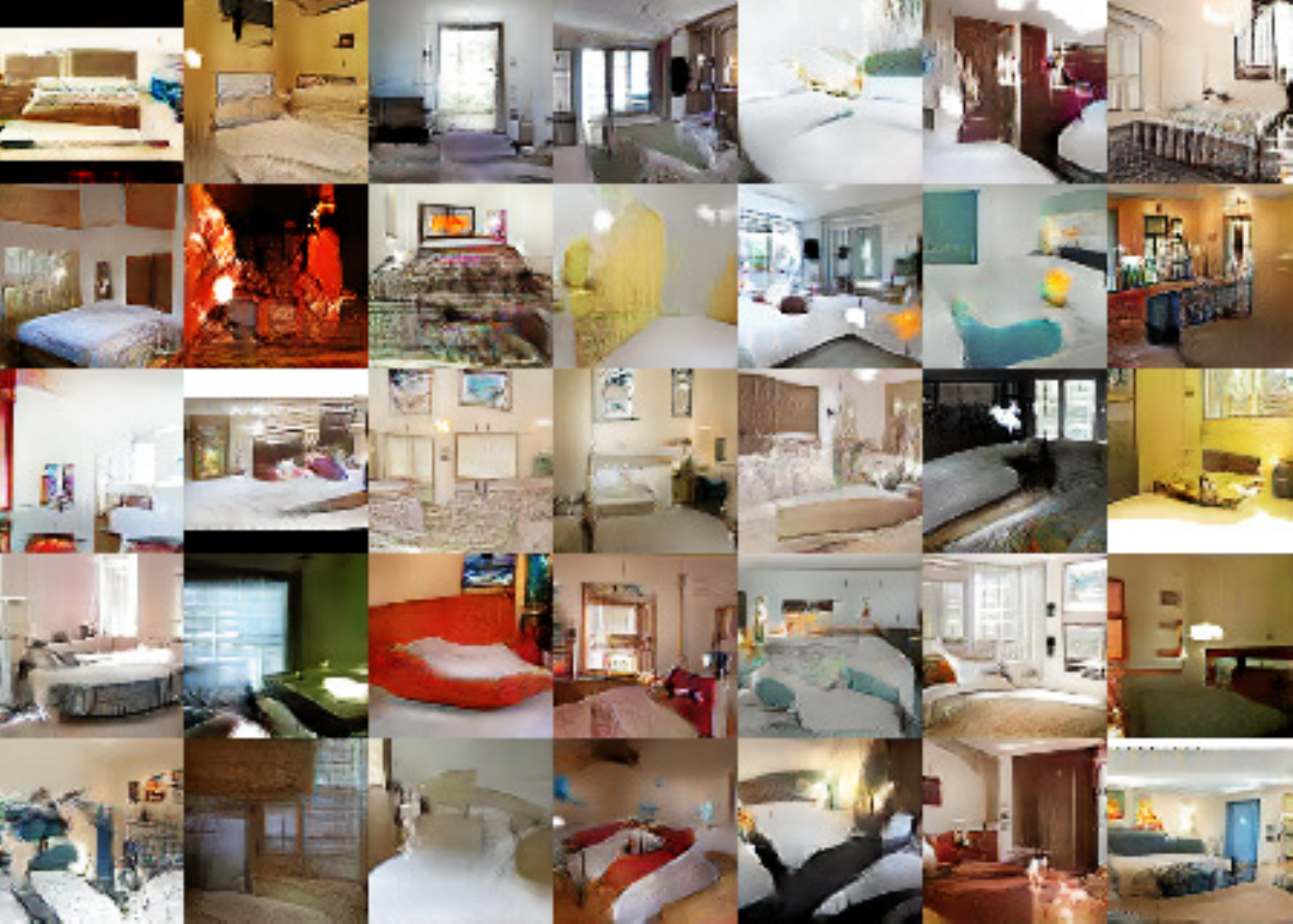} & 
\includegraphics[width=0.3\textwidth]{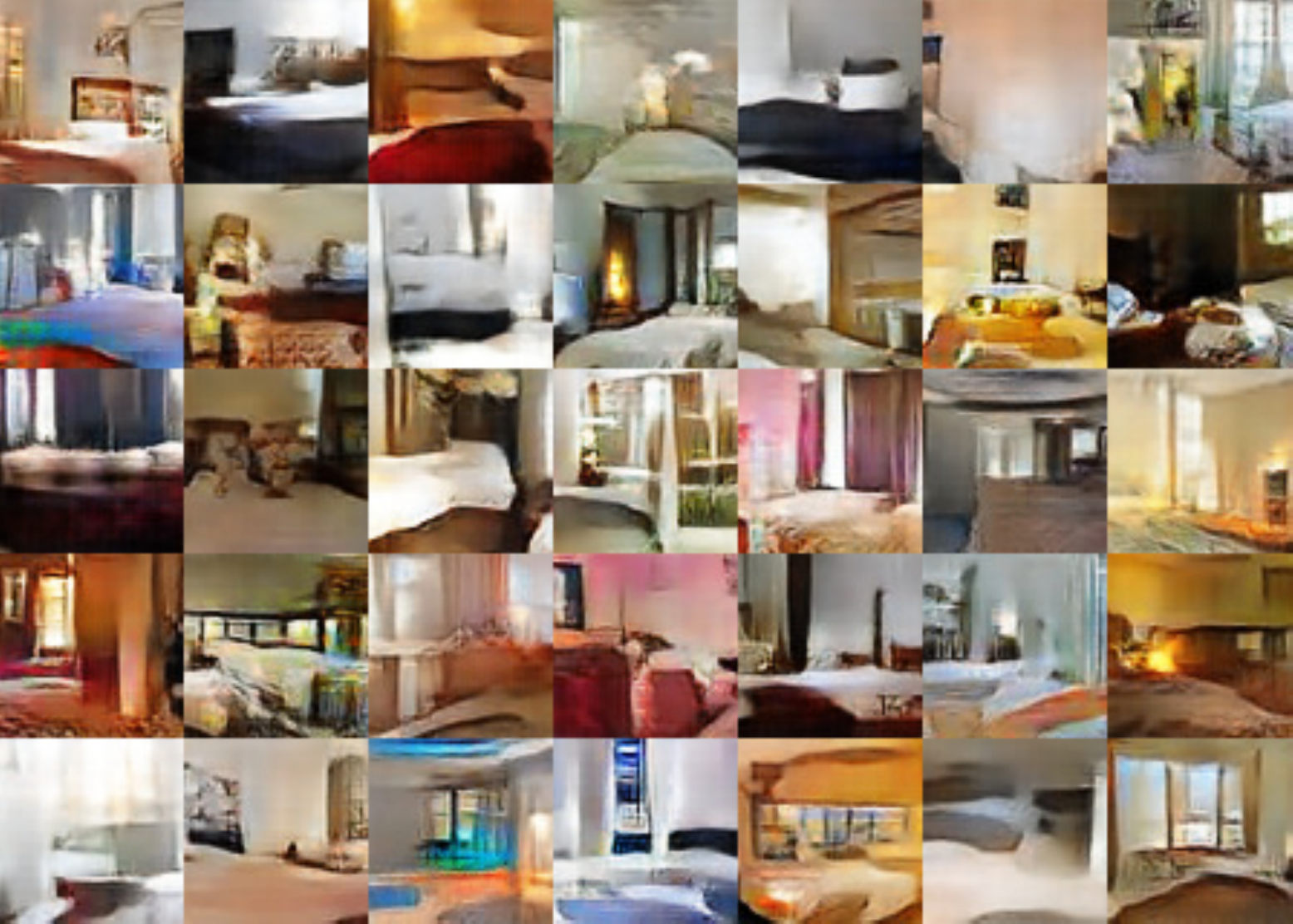} \\
\fttmp  DCGAN & \fttmp  SteinGAN\\
\end{tabular}
\caption{\fttmp  Images generated by DCGAN and our SteinGAN on LSUN.}
\label{fig:room}
\end{figure*}

\subsection{Deep Autoencoder-based SteinGAN}

\newcommand{\enc}{\mathrm{E}}
\newcommand{\dec}{\mathrm{D}}
In order to obtain better results on realistic datasets,  we test a more complex energy model based on deep autoencoder: 
\begin{align}\label{equ:px}
p(x|\theta) \propto \exp(-  || x -   \dec(\enc(x;~ \theta);~\theta) ||), 
\end{align}
where $x$ denotes the image and $\enc(\cdot)$, $\dec(\cdot)$ is a pair of encoder/decoder function, indexed by parameter $\theta$. 
This choice is motivated by Energy-based GAN \citep{zhao2016energy} in which the autoencoder is used as a discriminator but without a probabilistic interpretation. 
We assume $G(\xi;~\eta)$ to be a neural network whose input $\xi$ is a $100$-dimensional random vector drawn by $\mathrm{Uniform}([-1,1])$. 
Leveraging the latent representation of the autoencoder, 
we find it useful to define the kernel $k(x,x')$ on the encoder function, that is, 
$
k(x, x') = \exp(-\frac{1}{h^2} ||\enc(x;~\theta) - \enc(x';~\theta)||^2). 
$
We take the bandwidth to be $h=0.5 \times \mathrm{med} /\log m$, where $\mathrm{med}$ is the median of the pairwise distances between $\enc(x)$ on the image simulated by $f(\eta;~ \xi)$. 
This makes the kernel change adaptively based on both $\theta$ (through $\E(x;~\theta)$) and $\eta$ (through bandwidth $h$). 
Note that the theory of SVGD does not put constraints on the choice of positive definite kernels,  
and it allows us to obtain better results by changing the kernel adaptively during the algorithm. 

Some datasets include both images $x$ and 
their associated discrete labels $y$. In these cases, we train a joint energy model on $(x,y)$ 
to capture both the inner structure of the images and its predictive relation with the label,  which allows us to simulate images 
with a control on the category which it belongs to. Our joint energy model is defined by  
\begin{align}\label{equ:pxy}
p(x, y|\theta) \propto \exp\big \{&-  || x -   \dec(\enc(x;~\theta);~\theta) ||  \nonumber\\
&- \max[m,~\sigma(y, ~ \enc(x;~\theta))]\big \},  
\end{align}
where $\sigma(\cdot,\cdot)$ is the cross entropy loss function of a fully connected output layer. 
In this case, our neural sampler first draws a label $y$ randomly according to the empirical counts in the dataset, 
and then passes $y$ into a neural network together with a $100\times 1$ random vector $\xi$ to generate image $x$. 
This allows us to generate images for particular categories by controlling the value of input $y$. 

We compare our algorithm with DCGAN. We use the same generator architecture as DCGAN.
To be fair, our energy model has comparable or less parameters than the discriminator in the DCGAN. 
The number of parameters used are summarized in Table \ref{tbl:parameter}.

\paragraph{Stabilization}
In practice, we find it is useful to modify \eqref{steindd} in Algorithm~\ref{alg:steingan}  to be 
\begin{align*}
\theta \gets \theta  + \frac{\mu}{m}\sum_{i=1}^m(\nabla_\theta f(x_i^+; \theta) -  (1-\gamma) \nabla_\theta f(x_i^-; \theta)),
\end{align*}
where $\gamma$ is a discount factor (which we take to be $\gamma = 0.7$). 
This is equivalent to maximizing a regularized likelihood: 
$$\max_\theta  \{ \log   p(x |\theta)  +  \gamma \Phi(\theta)\}.$$
where $\Phi(\theta) = \log Z(\theta)$ is the log-partition function (see \eqref{eq:energy}); note that $\exp( \gamma \Phi(\theta))$ is a conjugate prior of $p(x|\theta)$. 

We initialize the weights of both the generator and discriminator from Gaussian distribution $\mathcal{N}(0,0.02)$, 
and train them using Adam \citep{kingma2014adam} with a learning rate of $0.001$ for the generator and $0.0005$ for the energy model (the discriminator).  
To keep the generator and discriminator approximately aligned during training, 
we speed up the $\theta$-update (by increasing the discount factor to $0.9$) when the energy of the real data batch is larger than the energy of the simulated images. 
We used the architecture guidelines for stable training suggested in DCGAN \citep{radford2015unsupervised}.

\begin{table}[t]
\centering
\begin{tabular}{ccc}
\hline
& \fttmp Cifar10 & \fttmp CelebA \& \fttmp LSUN \\
\hline
\fttmp DCGAN & $\fttmp \sim17m$ & $\fttmp \sim17m$ \\
\fttmp SteinGAN & $\fttmp \sim10m$ & $\fttmp \sim2.5m$ \\
\hline
\end{tabular}
\caption{\small Comparison of number of parameters used in discriminator networks. We use the same generator network as DCGAN through out our experiments.}
\label{tbl:parameter}
\end{table}

\paragraph{Discussion}
CIFAR-10 includes diverse objects, but has only 50,000 training examples.
Figure~\ref{fig:cifar10} shows examples of simulated images by DCGAN and SteinGAN generated conditional on each category, which look equally well visually. 
It is still an open question on how to quantitively evaluate the quantities of simulated images. 
In order to gain some understanding, here we report two different scores, including the inception score proposed by \citet{salimans2016improved}, and  
the classification accuracy when training ResNet using $50,000$ simulated images as train sets, evaluated on a separate held-out testing set never seen by the GAN models.  
Besides DCGAN and SteinGAN, we also evaluate another simple baseline obtained by subsampling 500 real images from the training set and duplicating them 100 times. 
We observe that these scores capture rather different perspectives of image generation:
the inception score favors images that look realistic individually and have uniformly distributed labels; as a result, 
the inception score of the duplicated 500 images is almost as high as the real training set. 
We find that the inception score of SteinGAN is comparable with DCGAN. 
On the other hand, the classification accuracy measures the amount information (in terms of classification using ResNet) captured in the simulated image sets;  
we find that SteinGAN achieves higher classification accuracy, suggesting that it captures, at least in some perspective, more information in the training set. 

Figure~\ref{fig:face} and \ref{fig:room} visualize the results on CelebA (with more than 200k face images) and LSUN (with nearly 3M bedroom images), respectively. 
We cropped and resized both dataset images into $64\times 64$.

\section{Conclusion}\label{sec:conclusion}
We propose a number of new algorithms for learning 
deep energy models, and demonstrate their properties. 
We show that our SteinCD performs well in term of test likelihood, 
while SteinGAN performs well in terms of generating realistic looking images. 
Our results suggest promising directions 
for learning better models by combining GAN-style methods with traditional energy-based learning. 

\bibliography{icml}

\begin{thebibliography}{32}
\providecommand{\natexlab}[1]{#1}
\providecommand{\url}[1]{\texttt{#1}}
\expandafter\ifx\csname urlstyle\endcsname\relax
  \providecommand{\doi}[1]{doi: #1}\else
  \providecommand{\doi}{doi: \begingroup \urlstyle{rm}\Url}\fi

\bibitem[Andrychowicz et~al.(2016)Andrychowicz, Denil, Gomez, Hoffman, Pfau,
  Schaul, and de~Freitas]{andrychowicz2016learning}
Andrychowicz, Marcin, Denil, Misha, Gomez, Sergio, Hoffman, Matthew~W, Pfau,
  David, Schaul, Tom, and de~Freitas, Nando.
\newblock Learning to learn by gradient descent by gradient descent.
\newblock \emph{arXiv preprint arXiv:1606.04474}, 2016.

\bibitem[Arjovsky et~al.(2017)Arjovsky, Chintala, and Bottou]{wgan}
Arjovsky, Martin, Chintala, Soumith, and Bottou, L{\'e}on.
\newblock Wasserstein gan.
\newblock \emph{arXiv preprint arXiv:1701.07875}, 2017.

\bibitem[Carreira-Perpinan \& Hinton(2005)Carreira-Perpinan and
  Hinton]{carreira2005contrastive}
Carreira-Perpinan, Miguel~A and Hinton, Geoffrey~E.
\newblock On contrastive divergence learning.
\newblock In \emph{AISTATS}, volume~10, pp.\  33--40. Citeseer, 2005.

\bibitem[Cover \& Thomas(2012)Cover and Thomas]{cover2012elements}
Cover, Thomas~M and Thomas, Joy~A.
\newblock \emph{Elements of information theory}.
\newblock John Wiley \& Sons, 2012.

\bibitem[Dziugaite et~al.(2015)Dziugaite, Roy, and
  Ghahramani]{dziugaite2015training}
Dziugaite, Gintare~Karolina, Roy, Daniel~M., and Ghahramani, Zoubin.
\newblock Training generative neural networks via maximum mean discrepancy
  optimization.
\newblock In \emph{Conference on Uncertainty in Artificial Intelligence (UAI)},
  2015.

\bibitem[Geyer(1991)]{geyer1991markov}
Geyer, Charles~J.
\newblock Markov chain monte carlo maximum likelihood.
\newblock 1991.

\bibitem[Goodfellow et~al.(2014)Goodfellow, Pouget-Abadie, Mirza, Xu,
  Warde-Farley, Ozair, Courville, and Bengio]{goodfellow2014generative}
Goodfellow, Ian, Pouget-Abadie, Jean, Mirza, Mehdi, Xu, Bing, Warde-Farley,
  David, Ozair, Sherjil, Courville, Aaron, and Bengio, Yoshua.
\newblock Generative adversarial nets.
\newblock In \emph{Advances in Neural Information Processing Systems}, pp.\
  2672--2680, 2014.

\bibitem[Goodfellow et~al.(2016)Goodfellow, Bengio, and
  Courville]{goodfellow2016deep}
Goodfellow, Ian, Bengio, Yoshua, and Courville, Aaron.
\newblock \emph{Deep learning}.
\newblock MIT Press, 2016.

\bibitem[He et~al.(2016)He, Zhang, Ren, and Sun]{he2016deep}
He, Kaiming, Zhang, Xiangyu, Ren, Shaoqing, and Sun, Jian.
\newblock Deep residual learning for image recognition.
\newblock In \emph{Proceedings of the IEEE Conference on Computer Vision and
  Pattern Recognition}, pp.\  770--778, 2016.

\bibitem[Hinton(2002)]{hinton2002training}
Hinton, Geoffrey~E.
\newblock Training products of experts by minimizing contrastive divergence.
\newblock \emph{Neural computation}, 14\penalty0 (8):\penalty0 1771--1800,
  2002.

\bibitem[Hinton \& Salakhutdinov(2006)Hinton and
  Salakhutdinov]{hinton2006reducing}
Hinton, Geoffrey~E and Salakhutdinov, Ruslan~R.
\newblock Reducing the dimensionality of data with neural networks.
\newblock \emph{science}, 313\penalty0 (5786):\penalty0 504--507, 2006.

\bibitem[Hyv{\"a}rinen(2005)]{hyvarinen2005estimation}
Hyv{\"a}rinen, Aapo.
\newblock Estimation of non-normalized statistical models by score matching.
\newblock In \emph{Journal of Machine Learning Research}, pp.\  695--709, 2005.

\bibitem[Kim \& Bengio(2016)Kim and Bengio]{kim2016deep}
Kim, Taesup and Bengio, Yoshua.
\newblock Deep directed generative models with energy-based probability
  estimation.
\newblock \emph{arXiv preprint arXiv:1606.03439}, 2016.

\bibitem[Kingma \& Ba(2014)Kingma and Ba]{kingma2014adam}
Kingma, Diederik and Ba, Jimmy.
\newblock Adam: A method for stochastic optimization.
\newblock \emph{arXiv preprint arXiv:1412.6980}, 2014.

\bibitem[Koller \& Friedman(2009)Koller and Friedman]{koller2009probabilistic}
Koller, Daphne and Friedman, Nir.
\newblock \emph{Probabilistic graphical models: principles and techniques}.
\newblock MIT press, 2009.

\bibitem[LeCun et~al.(2006)LeCun, Chopra, Hadsell, Ranzato, and
  Huang]{lecun2006tutorial}
LeCun, Yann, Chopra, Sumit, Hadsell, Raia, Ranzato, M, and Huang, F.
\newblock A tutorial on energy-based learning.
\newblock \emph{Predicting structured data}, 1:\penalty0 0, 2006.

\bibitem[Li et~al.(2015)Li, Swersky, and Zemel]{li2015generative}
Li, Yujia, Swersky, Kevin, and Zemel, Richard.
\newblock Generative moment matching networks.
\newblock In \emph{International Conference on Machine Learning}, pp.\
  1718--1727, 2015.

\bibitem[Liu \& Wang(2016)Liu and Wang]{liu2016stein}
Liu, Qiang and Wang, Dilin.
\newblock Stein variational gradient descent: A general purpose bayesian
  inference algorithm.
\newblock In \emph{Advances In Neural Information Processing Systems}, pp.\
  2370--2378, 2016.

\bibitem[Liu et~al.(2016)Liu, Lee, and Jordan]{liu2016kernelized}
Liu, Qiang, Lee, Jason~D, and Jordan, Michael~I.
\newblock A kernelized {Stein} discrepancy for goodness-of-fit tests.
\newblock In \emph{Proceedings of the International Conference on Machine
  Learning (ICML)}, 2016.

\bibitem[Liu et~al.(2015)Liu, Luo, Wang, and Tang]{liu2015faceattributes}
Liu, Ziwei, Luo, Ping, Wang, Xiaogang, and Tang, Xiaoou.
\newblock Deep learning face attributes in the wild.
\newblock In \emph{Proceedings of International Conference on Computer Vision
  (ICCV)}, 2015.

\bibitem[Ngiam et~al.(2011)Ngiam, Chen, Koh, and Ng]{ngiam2011learning}
Ngiam, Jiquan, Chen, Zhenghao, Koh, Pang~W, and Ng, Andrew~Y.
\newblock Learning deep energy models.
\newblock In \emph{Proceedings of the 28th International Conference on Machine
  Learning (ICML-11)}, pp.\  1105--1112, 2011.

\bibitem[Nowozin et~al.(2016)Nowozin, Cseke, and Tomioka]{nowozin2016f}
Nowozin, Sebastian, Cseke, Botond, and Tomioka, Ryota.
\newblock f-gan: Training generative neural samplers using variational
  divergence minimization.
\newblock \emph{arXiv preprint arXiv:1606.00709}, 2016.

\bibitem[Radford et~al.(2015)Radford, Metz, and
  Chintala]{radford2015unsupervised}
Radford, Alec, Metz, Luke, and Chintala, Soumith.
\newblock Unsupervised representation learning with deep convolutional
  generative adversarial networks.
\newblock \emph{arXiv preprint arXiv:1511.06434}, 2015.

\bibitem[Salimans et~al.(2016)Salimans, Goodfellow, Zaremba, Cheung, Radford,
  and Chen]{salimans2016improved}
Salimans, Tim, Goodfellow, Ian, Zaremba, Wojciech, Cheung, Vicki, Radford,
  Alec, and Chen, Xi.
\newblock Improved techniques for training gans.
\newblock In \emph{Advances in Neural Information Processing Systems}, pp.\
  2226--2234, 2016.

\bibitem[Scholkopf \& Smola(2001)Scholkopf and Smola]{scholkopf2001learning}
Scholkopf, Bernhard and Smola, Alexander~J.
\newblock \emph{Learning with kernels: support vector machines, regularization,
  optimization, and beyond}.
\newblock MIT press, 2001.

\bibitem[Snijders(2002)]{snijders2002markov}
Snijders, Tom~AB.
\newblock Markov chain monte carlo estimation of exponential random graph
  models.
\newblock \emph{Journal of Social Structure}, 3\penalty0 (2):\penalty0 1--40,
  2002.

\bibitem[Wang \& Liu(2016)Wang and Liu]{wang2016steingan}
Wang, Dilin and Liu, Qiang.
\newblock Learning to draw samples: With application to amortized mle for
  generative adversarial learning.
\newblock \emph{arXiv preprint arXiv:1611.01722}, 2016.

\bibitem[Welling(2009)]{welling2009herding}
Welling, Max.
\newblock Herding dynamical weights to learn.
\newblock In \emph{Proceedings of the 26th Annual International Conference on
  Machine Learning}, pp.\  1121--1128. ACM, 2009.

\bibitem[Xie et~al.(2016)Xie, Lu, Zhu, and Wu]{xie2016theory}
Xie, Jianwen, Lu, Yang, Zhu, Song-Chun, and Wu, Ying~Nian.
\newblock A theory of generative convnet.
\newblock \emph{arXiv preprint arXiv:1602.03264}, 2016.

\bibitem[Yu et~al.(2015)Yu, Seff, Zhang, Song, Funkhouser, and
  Xiao]{yu2015lsun}
Yu, Fisher, Seff, Ari, Zhang, Yinda, Song, Shuran, Funkhouser, Thomas, and
  Xiao, Jianxiong.
\newblock Lsun: Construction of a large-scale image dataset using deep learning
  with humans in the loop.
\newblock \emph{arXiv preprint arXiv:1506.03365}, 2015.

\bibitem[Zhai et~al.(2016)Zhai, Cheng, Feris, and Zhang]{zhai2016generative}
Zhai, Shuangfei, Cheng, Yu, Feris, Rogerio, and Zhang, Zhongfei.
\newblock Generative adversarial networks as variational training of energy
  based models.
\newblock \emph{arXiv preprint arXiv:1611.01799}, 2016.

\bibitem[Zhao et~al.(2016)Zhao, Mathieu, and LeCun]{zhao2016energy}
Zhao, Junbo, Mathieu, Michael, and LeCun, Yann.
\newblock Energy-based generative adversarial network.
\newblock \emph{arXiv preprint arXiv:1609.03126}, 2016.

\end{thebibliography}
\bibliographystyle{icml2017}

\end{document}